\documentclass[conference]{IEEEtran}
\pdfoutput=1
\IEEEoverridecommandlockouts
\usepackage{cite}
\usepackage{amsmath,amssymb,amsfonts}
\usepackage{algorithmic}
\usepackage{graphicx}
\usepackage{textcomp}
\usepackage{xcolor}
\usepackage{float}
\usepackage{graphics} 
\usepackage{epsfig} 
\usepackage{mathptmx} 
\usepackage{times} 
\usepackage{amsmath} 
\usepackage{amssymb}  
\usepackage{url}
\usepackage{color}
\usepackage{graphicx}
\usepackage{subcaption}
\usepackage{caption}
\usepackage{multirow}
\usepackage{url}
\usepackage{float}
\usepackage{array}
\usepackage{relsize}
\usepackage{sidecap}   
\usepackage{mwe} 
\usepackage{cite}
\usepackage{tikz}
\usepackage{textcomp}
\usepackage{hyperref}
\usepackage[ruled,linesnumbered]{algorithm2e}

\newcommand\copyrighttext{
	\footnotesize \textcopyright 2024 IEEE. Personal use of this material is permitted.  Permission from IEEE must be obtained for all other uses, in any current or future media, including reprinting/republishing this material for advertising or promotional purposes, creating new collective works, for resale or redistribution to servers or lists, or reuse of any copyrighted component of this work in other works.
	DOI: 10.23919/FUSION59988.2024.10706281}
\newcommand\copyrightnotice{
	\begin{tikzpicture}[remember picture,overlay]
		\node[anchor=south,yshift=10pt] at (current page.south) {\fbox{\parbox{\dimexpr\textwidth-\fboxsep-\fboxrule\relax}{\copyrighttext}}};
	\end{tikzpicture}
}

\definecolor{darkgreen}{HTML}{006400}
\definecolor{lightviolet}{HTML}{EE82EE}
\definecolor{copper}{HTML}{B87333}
\definecolor{blueviolet}{HTML}{8A2BE2}

\SetCommentSty{mycommfont}

\def\BibTeX{{\rm B\kern-.05em{\sc i\kern-.025em b}\kern-.08em
    T\kern-.1667em\lower.7ex\hbox{E}\kern-.125emX}}
\begin{document}

\title{
	Incorporating Navigation Context into Inland Vessel Trajectory Prediction: A Gaussian Mixture Model and Transformer Approach 
\thanks{This publication originates from a joint research project between the German Federal Waterways Engineering and Research Institute (BAW) and the Institute for Ship Technology, Ocean Engineering and Transport Systems (ISMT) (University of Duisburg-Essen), and was developed and written in collaboration with the Chair of Dynamics and Control (University of Duisburg-Essen).}
}

\author{\IEEEauthorblockN{Kathrin Donandt}
\IEEEauthorblockA{\textit{University of Duisburg-Essen} \\
\textit{Federal Waterways Engineering and Research Institute}\\
Karlsruhe, Germany\\
kathrin.donandt@baw.de}
\and
\IEEEauthorblockN{Dirk Söffker}
\IEEEauthorblockA{\textit{University of Duisburg-Essen} \\
Duisburg, Germany \\
soeffker@uni-due.de}
}

\maketitle
\copyrightnotice

\begin{abstract}
Using data sources beyond the Automatic Identification System (AIS) to represent the context in which a vessel is navigating and consequently improve situation awareness is still rare in machine learning approaches to vessel trajectory prediction (VTP). In inland shipping, where vessel movement is constrained within fairways, supplementary navigational context information is indispensable. In this contribution targeting inland VTP, Gaussian Mixture Models are applied, on a fused dataset of AIS and discharge measurements, to generate multi-modal distribution curves, capturing typical lateral vessel positioning in the fairway and dislocation speeds along the waterway. 
By subsequently sampling the probability density curves of the GMMs, feature vectors are derived which are used, together with spatio-temporal vessel features and fairway geometries, as input to a VTP transformer model. 
The incorporation of these distribution features of both the current and forthcoming navigation context improves prediction accuracy. The superiority of the model over a previously proposed navigation context-sensitive transformer model for inland VTP is shown. 
The novelty lies in the provision of preprocessed, statistics-based features representing the conditioned spatial context, rather than relying on the model to extract relevant features for the VTP task from contextual data.  
Oversimplification of the complexity of inland navigation patterns by assuming a single typical route or selecting specific clusters prior to model application is avoided by giving the model access to the entire distribution information. 
The methodology's generalizability is demonstrated through the usage of data of three distinct river sections. It can be integrated into an interaction-aware prediction framework, where insights into the positioning of the actual vessel behavior in the overall distribution at the current location and discharge can further enhance trajectory prediction accuracy.

\end{abstract}


\section{Introduction}
Predicting vessel trajectories is a pivotal aspect in ensuring the safety and efficiency of autonomous shipping operations. For an autonomous vessel to navigate without collision and effectively, it must possess a comprehensive understanding of the likely future paths of nearby vessels, enabling it to determine the most advantageous course. Anticipating potential collisions allows for timely avoidance maneuvers, while strategic trajectory planning optimizes energy consumption. In the context of inland shipping, characterized by confined navigation channels, fluctuating flow velocities, and shallow water conditions, proactive planning is imperative. This requires a precise assessment of the current navigation scenario and its possible developments. The trajectory prediction is essential for evaluating the options. 
Inspired by the success seen in other fields, particularly in the realm of automotive trajectory prediction, numerous approaches to vessel trajectory prediction (VTP) in the maritime domain have recently embraced deep learning (DL) techniques, as evidenced by works like \cite{Yang.2024,Li.2023a,Zhang.2022}. 
The efficacy of these prediction models depends largely on the quality and comprehensiveness of the training data. Many DL-based studies on VTP rely solely on AIS data, forecasting future trajectories based on observed spatio-temporal displacement features such as latitude, longitude, speed over ground (SOG), and course over ground (COG). These approaches prove inadequate for inland navigation due to the unknown spatial constraints. 
Consequently, this study proposes the inclusion of supplementary features that encapsulate spatial conditions which are derived from statistical pre-evaluations of AIS data annotated with hydrological measurements and basic river geometries. The proposed model is provided with ``statistical'' maps illustrating the discharge-dependent and local distribution of vessel locations in relation to the fairway and of vessel dislocation speeds. 
By employing Gaussian Mixture Models (GMM) to model these distributions, their  inherently multi-modal nature is effectively considered. 
Aside from solely focusing on individual VTP in this study, disregarding the influence of surrouding ships, the integration of features representing navigational behavior distributions is anticipated to offer significant advantages, particularly in the context of joint VTP. The DL-based VTP research is advancing to approaches considering interacting vessels \cite{Feng.2022,Sekhon.2020,Liu.2022}, which is especially crucial for the inland navigation domain.  
By incorporating knowledge of how neighboring vessels behave in relation to the expected behavior derived from distribution-representing features, an enhancement in trajectory prediction performance is expected. 

The paper is organized as follows: In Section \ref{sec:RelatedWork} an overview of DL-based VTP methods is provided, emphasizing the use of multi-source datasets, the transformer architecture, and the integration of GMMs, contextualizing this contribution within this landscape. In Section \ref{sec:Method}, the proposed methodology, including the generation of supplementary features using GMMs and the transformer-based VTP model, is detailed. In Section \ref{sec:Data}, the data sources and their fusion are introduced. The evaluation of the proposed model, including an ablation study and comparison with baselines and a previously proposed transformer-based VTP model, is presented in Section \ref{sec:Results}. Finally, the paper concludes with closing remarks, highlighting limitations and suggesting future directions.

\section{Related work}\label{sec:RelatedWork}
\subsection{Multi-source datasets in DL-based VTP}\label{sec:multisource}
Few previous DL methods for VTP include supplementary data in addition to AIS.  Dijt \& Mettes \cite{DijtPimandMettesPascal.2020} propose a hybrid DL architecture for VTP of inland vessels. Apart from AIS data, the authors use data from radar images
and electronic navigational charts (ENCs). A CNN sub-module learns, as a side task, to semantically segment radar images by mapping them to ENCs, and the output of the convolutional layer is used in addition to the SOG and COG values as input to an RNN. 
Mehri et al. \cite{Mehri.2021} use contextual data in the training data setup step. For the LSTM model, trained to correct the prediction of a physical model for motion prediction, compressed trajectories from AIS with reduced redundancies are used as data source. For a logical consistent compression, shoreline data is considered. Shiptype-specifically trained LSTM models are then applied whenever a prediction of the physical model is illogical, i.e. is crossing land or untraversable corridores.
Huang et al. \cite{Huang.2022Triple} fuse AIS and discretized meterological data including wind, wave, ocean current, and ocean temperature information, and train a VTP model combining convolutional neural networks with multi-head attention mechanism.  
Finally, Donandt et al. \cite{Donandt.2023} use fairway boundary data and waterway kilometer (KM) information to define the dislocation of inland vessels in relation to a river-specific coordinate system, and train a transformer encoder-decoder to predict future trajectories of inland vessels.

\subsection{Transformer models for VTP} 
In line with prior developments (\hspace{1sp}\cite{DijtPimandMettesPascal.2020,Capobianco.2021,Murray.2021,Sekhon.2020,Nguyen2018,You.2020,Forti.2020,Jurkus.2023}), VTP is modeled as a sequence-to-sequence task, employing an encoder-decoder DL architecture. Departing from the prevalent LSTM- or GRU-based encoder-decoder models, a transformer encoder-decoder model \cite{Vaswani.2017}, gaining attention in the VTP research recently \cite{Yang.2024}, is chosen instead due to its parallel computation and success  in modeling complex sequence-to-sequence tasks.  
One of the first works applying transformers to VTP is the TrAISformer \cite{Nguyen.2024}, first proposed in \cite{Nguyen2018}, where discretized latitude, longitude, SOG, and COG features are used in a classification framework. The future trajectory is predicted by iteratively feeding back the last prediction to the model, increasing the input sequence with each prediction time step. 
More recently, Huang et al. \cite{Huang.2022Triple} use solely the transformer encoder part (self-attention) to learn the dependency between the features obtained from convolutional modules processing the fused AIS and meterological data. 
The future trajectory is, however, generated by a further CNN module, with the transformer encoder serving only as a sub-module rather than the main VTP model. 
Donandt et al.\cite{Donandt.2022} introduce a context-sensitive classification transformer encoder-decoder, CSCT, which re-frames the regression as a classification task as proposed in \cite{Nguyen2018}. The authors use a vector describing the upcoming river context in terms of curvature for the initialization of the decoder to enable spatial situation awareness. 
Unlike \cite{Nguyen2018}, the entire output sequence is generated at once by the transformer decoder based on the encoded observation sequence, only considering previous predictions in the decoder itself through masking, as originally proposed in \cite{Vaswani.2017}.  
Finally, the recent work of Liu et al. \cite{Liu.2024} introduces a spatio-temporal multi-graph transformer network to simultaneously predict the trajectories of interacting vessels. Transformer components are used to encode the individual vessel trajectories and the graph representation of the traffic scene, while the prediction of the vessel trajectories are accomplished with an LSTM. 

\subsection{Gaussian Mixture Models in the context of DL-based VTP}
Gaussian Mixture Models are applied in Murray \& Perera \cite{Murray.2020} to cluster historical trajectories. 
Based on the cluster of trajectories to which a selected vessel's trajectory is assigned, a trajectory prediction is obtained via a dual linear autoencoder model. 
In contrast to the usage of GMM in a prior step to the VTP task, Dalsnes et al. \cite{Dalsnes.2018}, fit a GMM on the multiple possible trajectories obtained from the proposed  Neighbor Course Distribution Method (NCDM). Thus, a probabilistic model of the future positions is obtained, accounting for uncertainty and multi-modality.  
\begin{figure*}[hbt!]
	\begin{subfigure}{.5\textwidth}
		\includegraphics[trim=2 10 2 10, clip, width=\linewidth]{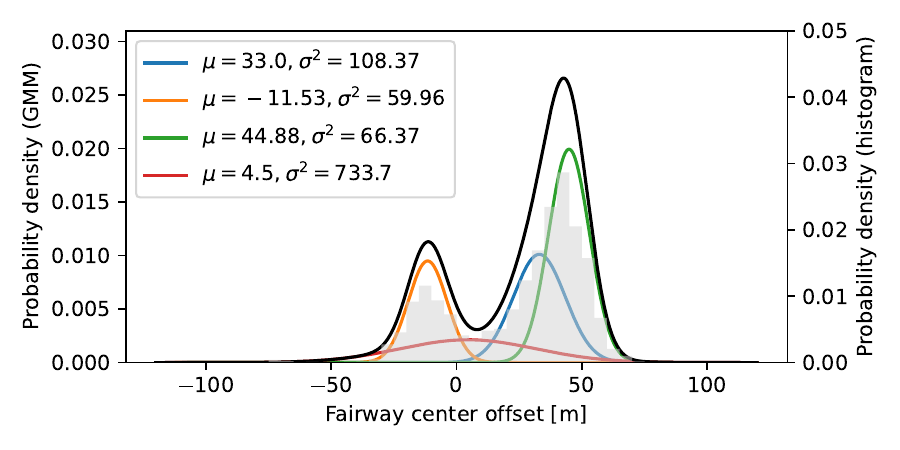}
		\caption{$Q = 1750$ {\upshape m³/s}}
		\label{fig:GMM1d1750}
	\end{subfigure}
	\begin{subfigure}{.5\textwidth}
		\includegraphics[trim=2 10 2 10, clip, width=\linewidth]{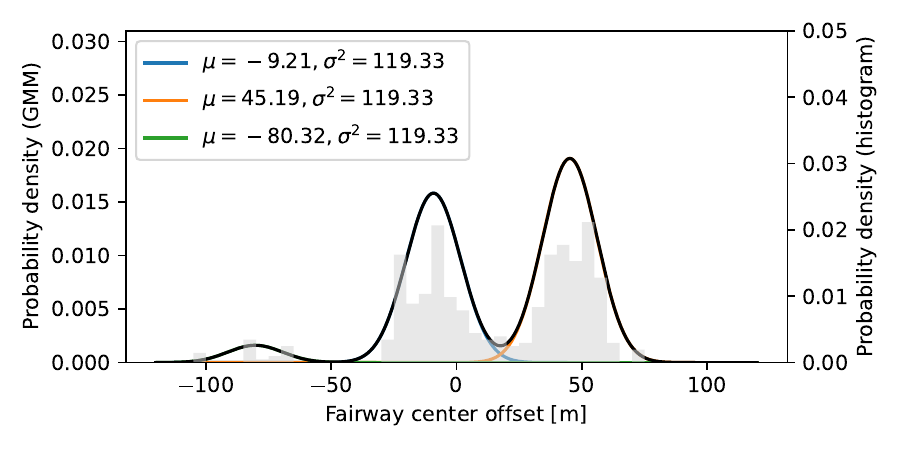}
		\caption{$Q = 3250$ {\upshape m³/s}}
		\label{fig:GMM1d3250}
	\end{subfigure}
	\caption{Example of lateral GMMs: Distribution of fairway center offsets of upstream navigating vessels at Rhine KM 598.6 at different discharges ($Q$).}\label{fig:GMM1d}
\end{figure*}
\hfill \break
This study presents an extension of previous approaches by integrating hydrological data, i.e. discharge, with AIS data to enhance the spatial situation awareness, particularly for the inland shipping domain. 
Different from earlier approaches delegating the extraction of pertinent features for VTP from contextual data sources such as maps to the model, the prior fusion of contextual data with vessel driving patterns is proposed. Unlike previous works using transformer components, a holistic approach is adopted by employing the entire transformer encoder-decoder model for the prediction of future time steps. Instead of the formerly used classification reframing in transformer-based approaches, the (dis-)location features are not discretized any more, resulting in a reduction of model parameters. The output consists of parameters of a bivariate Gaussian distribution, thus allowing approximation of the uncertainty \cite{Capobianco.2021}. 
Moreover, GMMs are employed to generate a kind of cluster information in the form of sampled probability density curves. By providing these samplings as input, the model is not constrained to a specific cluster.  
Rather, it incorporates the entire distribution information, ensuring a more comprehensive understanding of vessel behavior.
\section{Methodology}\label{sec:Method}
\begin{figure}[hbt]
	\begin{subfigure}{.475\textwidth}
		\includegraphics[trim=0 12 0 11,clip, width=\linewidth]{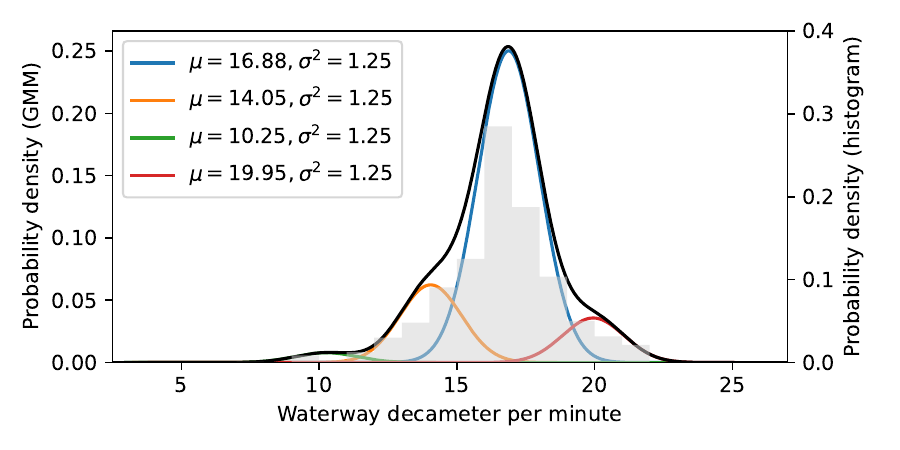}
		\caption{Lane 1}
		\label{fig:GMM2dSpur1}
	\end{subfigure}
	\begin{subfigure}{.475\textwidth}
		\includegraphics[trim=0 12 0 11,clip, width=\linewidth]{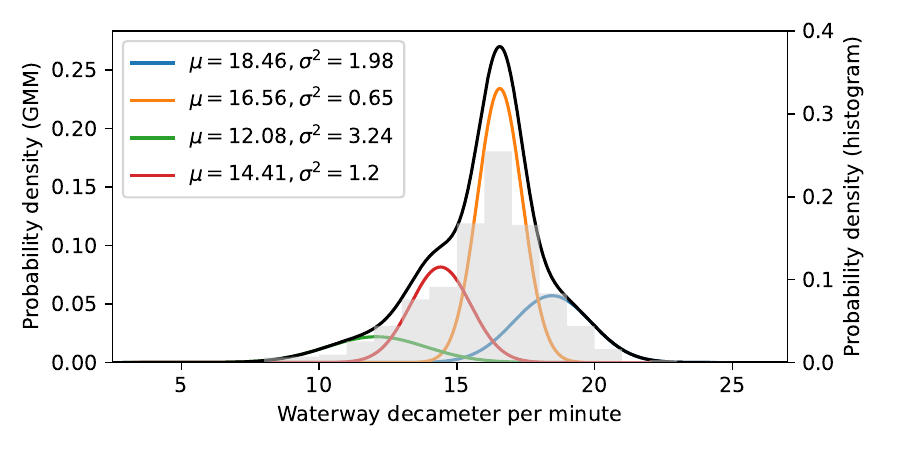}
		\caption{Lane 2}
		\label{fig:GMM2dSpur2}
	\end{subfigure}
	\begin{subfigure}{.475\textwidth}
		\includegraphics[trim=0 12 0 11,clip, width=\linewidth]{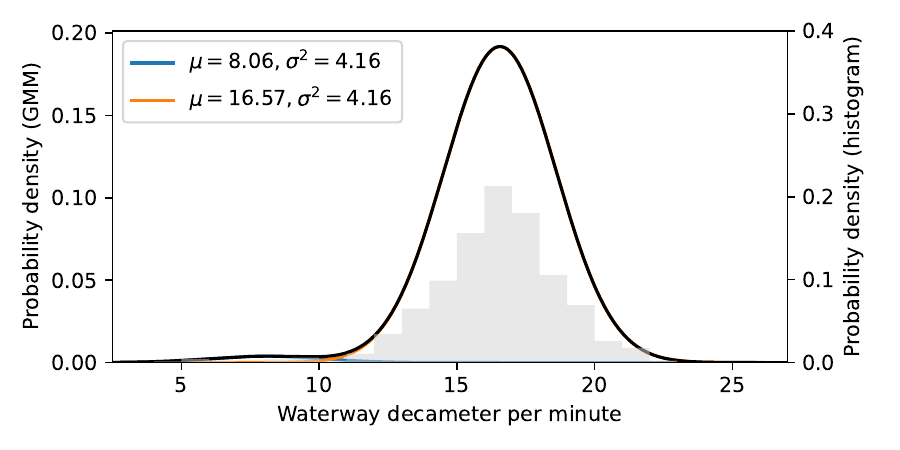}
		\caption{Lane 3}
		\label{fig:GMM2dSpur3}
	\end{subfigure}
	\begin{subfigure}{.475\textwidth}
		\includegraphics[trim=0 12 0 11,clip, width=\linewidth]{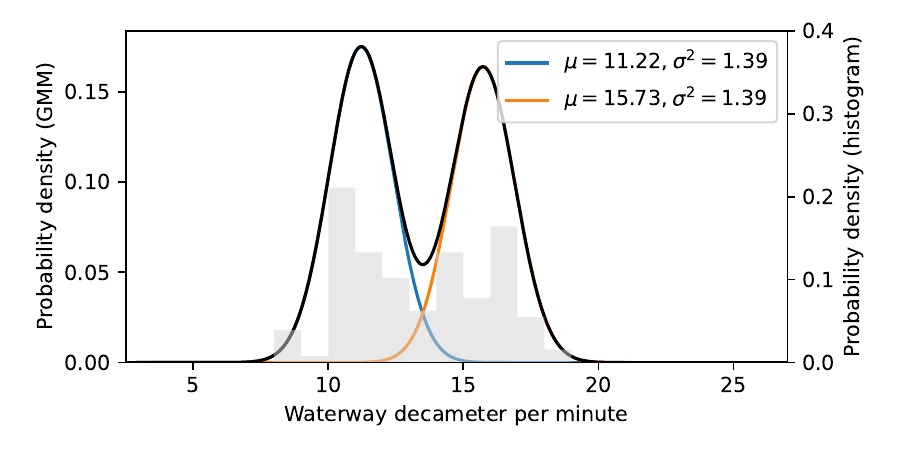}
		\caption{Lane 4}\label{fig:GMM2dSpur4}
	\end{subfigure}
	\setlength{\belowcaptionskip}{-8pt} 
	\caption{Examples of longitudinal GMMs: Distributions of KM distance (in decameters for better visibility) per minute at different locations in relation to the fairway at Rhine KM 608.4 at a discharge of 1000 \upshape{m³/s} (upstream navigation).}\label{fig:GMM2dSpur}
\end{figure}
\subsection{Statistical feature generation}\label{sec:stat} 
The driving behavior of inland vessels navigating free-flowing rivers is significantly affected by the discharge. Discharge $Q$ refers to the volume of water flowing through a cross-section of a waterway per unit of time. 
It is often given in cubic meters per second. 
Discharge variation result in modified flow velocities and river depths which affect the vessel's speed and maneuverability\cite{BAW.2016}. 
Incorporating discharge information into prediction models is therefore expected to enhance the accuracy of vessel trajectory predictions. 
Thus, the input data is augmented with feature vectors that characterize the discharge-dependent distribution of the lateral offsets from the fairway center and of the dislocation velocity in terms of waterway kilometer (KM) distance travelled per time interval - both at the vessel's current position as well as in the river section the vessel is heading to. 
Given the presumed multi-modal nature of both distributions, Gaussian Mixture Models (GMMs) are utilized to approximate them.  
A GMM can be employed to represent data assumed to originate from a combination of multiple Gaussian distributions. It is a parametric probability density function which returns the probability density by summing up the weighted densities of the Gaussian components. 
The estimation of the GMM parameters is typically accomplished through the Expectation-Maximization algorithm.
To achieve a GMM that best fits the data, two hyperparameters, the number of mixture components and covariance constraints (spherical, diagonal, tied, or full), can be determined using grid search. The model with the lowest Baysian Information Criterion (BIC) is selected. The BIC is defined as $b \times log(S) - 2log(\hat{L})$, 
where $b$ is the number of parameters of the GMM, $S$ is the dataset size, and $\hat{L}$ is the likelihood of the data under the given model. This evaluation criterion penalizes higher numbers of parameters to avoid overfitting, while maximizing the log-likelihood of the data. In this study, the maximum number of mixture components used in the grid search is set to 4 which is assumped to be sufficient for the distributions considered.  
For the representation of the fairway center offset distributions, GMMs are fitted on fairway center offsets of vessel positions from AIS. A separate GMM is generated for each navigation direction, waterway hectometer, and discharge. To avoid data scarcity issues, the KMs of the trajectories are rounded to full hectometers and the measured discharges to the nearest multiple of 250 m³/s. These GMMs are refered to as ``lateral GMMs'' in the subsequent sections. 
To model the distributions of KM distances travelled per minute, the waterway is devided into four artificial ``lanes'': two inner lanes, which cover the region between fairway center and right and left boundary, respectively, and two outer lanes, encompassing the remaining river area left and right of the fairway. A GMM is fitted on the KM distances per minute of the vessel trajectories from AIS - not only per navigation direction, waterway hectometer, and discharge  (rounded), but also per lane. These GMMs are termed ``longitudinal GMMs'' henceforth.
In Fig. \ref{fig:GMM1d1750} and \ref{fig:GMM1d3250}, depicting the probability density curves and components of lateral GMMs for the upstream direction at a selected Rhine KM, the impact of a discharge variation on the fairway center offsets is evident. 
At a lower discharge, the vessels typically navigate within 50 m to the left of the fairway center (negative offset), while at a higher discharge, they extend their navigation further to the left side. 
In Fig. \ref{fig:GMM2dSpur}, the four probability density curves and underlying components of the longitudinal GMMs per lane at a selected Rhine KM and discharge for upstream navigating vessels  are shown. From the far left (Fig. \ref{fig:GMM2dSpur1}) to the far right lane (Fig. \ref{fig:GMM2dSpur4}), a decrease of the most likely KM distances per minute is visible, and two clearly separable modes are identified by the GMM for the far right lane. Interestingly, shared variances are resulting in the lowest BIC scores for most GMMs depicted in Fig. \ref{fig:GMM1d} and Fig. \ref{fig:GMM2dSpur}. Once the optimal lateral GMMs are obtained for each navigation direction, hectometer, and discharge, bivariate spline interpolation is applied on sampling vectors of the lateral GMMs of each direction and discharge to obtain a function $gmm_{lat}^{dir,q}$, that, for a given navigation direction $dir$ and discharge $q$, returns the probability density value for any requested KM in the covered river section $k_{min}$ to $k_{max}$ and any covered fairway center offset in the range $-m_{max}$ to $m_{max}$. This interpolation is required as fitting a GMM for every potential KM (given with a precision of centimeters) for the vessel positions from AIS would be impractical due to the resulting data scarcity and computational demands. 
\begin{algorithm}
	\caption{Generation of lookup dictionary $L_{lat}$ for on-demand generation of vector $p_{lat}$}\label{alg:lookupdict}
	\SetKwInOut{Input}{Input}
	\SetKwInOut{Output}{Output}
	\SetKwFunction{Function}{Interpolate}
	\Input{Lateral GMMs for directions $\lbrace up, down \rbrace$, discharges $Q = \lbrace q_{min},q_{min}+250,...,q_{max}\rbrace$, and KMs $K = \lbrace k_{min}, k_{min}+0.1,...,k_{max}\rbrace$; offsets $M = \lbrace -m_{max}, -m_{max}+1,..., m_{max} \rbrace $.}
	\Output{Lookup dictionary $L_{lat}$ containing for each direction $dir$ and $q \in Q$ the function $gmm_{lat}^{dir,q}$.}
	\SetKwProg{Fn}{Function}{:}{}
	\Fn{\Function{lateral GMMs, $Q$, $K$, $M$}}
	{
		Let $L_{lat}$ be an empty dictionary; \\
		\ForEach{$dir \in \lbrace up, down\rbrace$}
		{
			\ForEach{$q \in Q$}
			{
				Let $A_{lat}[|K|][|M|]$ be an empty 2D array;\\
				$j=0$; \\
				\ForEach{$k \in K$}
				{
					$\hat{p}_{lat}^{(i)} = \sum_{c=1}^{C}{pdf_c(i-m_{max};\mu_c,\sigma_c^2)}$\tcp*[l]{Sample $\hat{p}_{lat}$ of length $|M|$ from the lateral GMM for $q$,$dir$, and $k$, with $C$ mixture components $ \mathcal{N}(\mu_c,\,\sigma_c^{2}$), $i\in \lbrace 0,2m_{max} \rbrace$.}
					$A_{lat}[j] = \hat{p}_{lat}$; \\
					$j = j+1$;
				}
				$gmm_{lat}^{dir,q} = BivariateSpline(K,M,A_{lat})$\tcp*[l]{Interpolate over $A_{lat}$, with coordinates $K$ and $M$.}
				$L_{lat}[(dir,q)] = gmm_{lat}^{dir,q}$;
		}}
		\KwRet{$L_{lat}$}
	}
\end{algorithm}
In Algorithm \ref{alg:lookupdict}, a lookup dictionary $L_{lat}$ for the considered river section, fairway center offset range, and discharge range $q_{min}$ to $q_{max}$ is constructed. This dictionary is subsequently employed to obtain on the fly the $D$-dimensional vector of probability densities, $p_{lat} = (d_{k_t}^1,...,d_{k_t}^D)^T$ for the KM $k_t$ of the vessel's position at time step $t$ and $D$ ordered and regularly spaced fairway center offsets. This vector constitutes a sampling vector of the probability density curve of the distribution of fairway center offsets. It is used as one of the additional inputs to the transformer model (Section \ref{sec:Model}) to represent the navigation context. The dimension $D$ is a hyperparamter and determined by $2m'r_d + 1$, where $m' \leq m_{max}$ is the maximum absolute fairway center offset chosen and $r_d$ is the selected resolution for the offset. 
For the longitudinal case, the procedure is identical, except for the additional consideration of the lane, thus resulting in functions designated as $gmm_{lon}^{dir,q,lane}$. The $V$-dimensional sampling vector $p_{lon} = (v_{k_t}^1,...,v_{k_t}^V)^T$ contains sampled densities of the probability density curve representing the distribution of KM distances per minute at the current KM $k_t$ in the lane the vessel is currently located in. Note that the entries both in $p_{lat}$ and $p_{lon}$ are obtained from interpolation and are thus approximate probability density values.
\begin{figure*}[hbt!]		
	\includegraphics[trim= 200 250 0 69, clip, width=\textwidth]{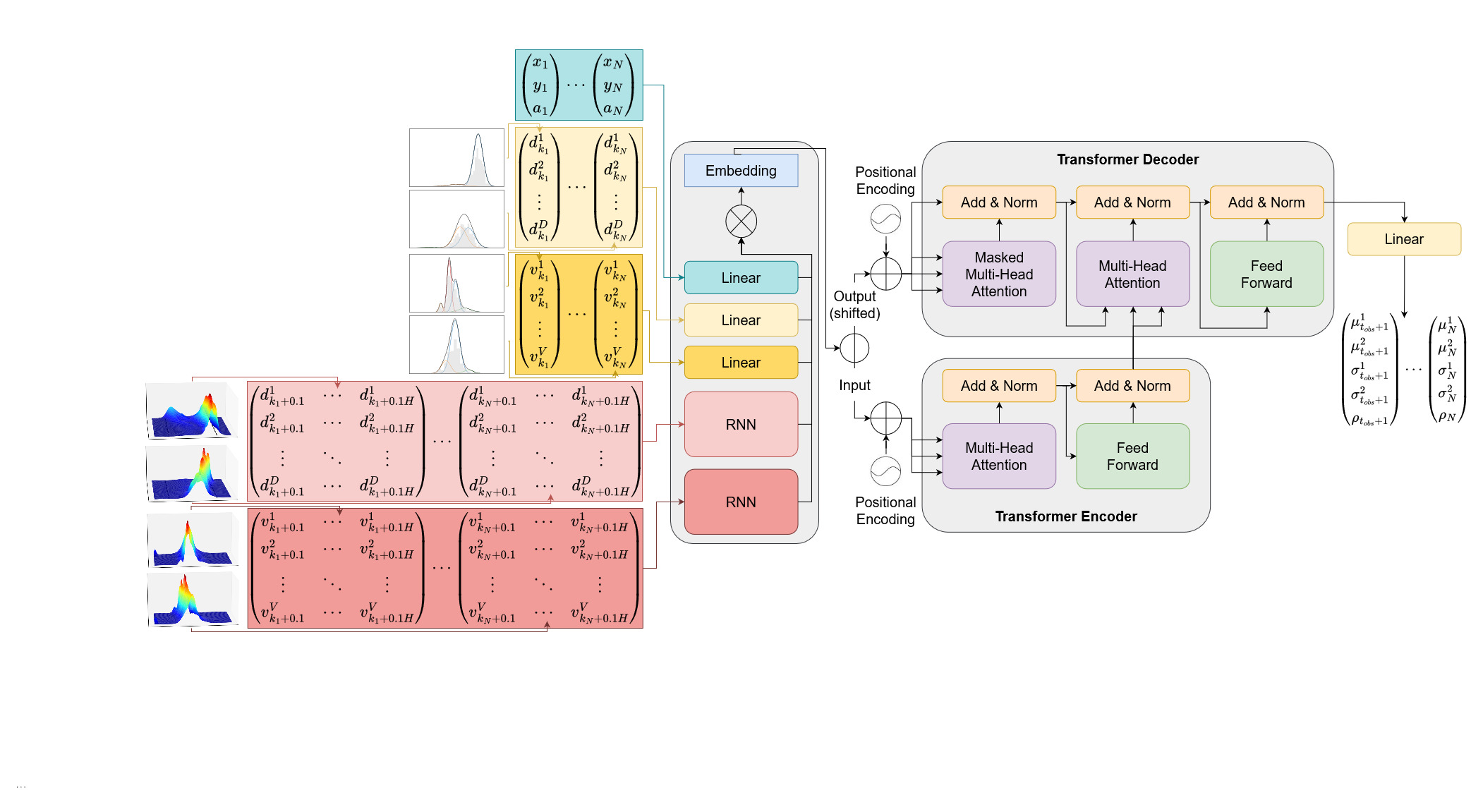}	
	\caption{Architecture of the proposed transformer model for inland vessel trajectory prediction.}
	\label{fig:architecture}
\end{figure*}
Analogously to $D$, $V$ is a hyperparameter, determined by $s_{max}r_v + 1$, where $s_{max}$ is the maximum KM distance per minute that is being considered, and $r_v$ the resolution. 
\subsection{Trajectory prediction model}\label{sec:Model} 
The trajectory prediction task is approached as a time series prediction problem, wherein the vessel's locations and additional contextual data within an observation time interval are used to predict the positions for a subsequent time interval. This is achieved through training a transformer encoder-decoder model to map the input sequence $i_1,...,i_{t_{obs}}$ to the output sequence $o_{t_{obs}+1},...,o_{N}$.
Specifically, the input time series element $i_t$ at time step $t \in \lbrace 1,...,t_{obs}\rbrace$ consists of three vectors and two matrices in this study, which is also visible on the left side of the model architecture diagram in Fig. \ref{fig:architecture}. The first vector is the vessel's (dis)location vector, $(x_t,y_t,a_t)^T$, where $x_t$, $y_t$, and $a_t$ are, respectively, the fairway center offset, the KM distance per minute, and the inverse of the fairway center radius, with different signs for right and left curves. 
The second vector is the $p_{lat}$ vector, which is obtained by looking up $gmm_{lat}^{dir,q}$ in $L_{lat}$ for the given navigation direction $dir$ and discharge $q$ and calling it on the current KM $k_t$ and the $D$ pre-defined fairway center offsets $(-m', -m'+r_d,...,m'-r_d, m')$. 
The third vector $p_{lon}$ is obtained analogously from  $gmm_{lon}^{dir,q,lane}(k_t,i)$, $\forall i \in  (0,0+r_v,...,v'-r_v,v')$ (the chosen $V$ KM distances per minute). 
The two matrices represent the lateral and longitudinal distribution information of the river section, the vessel is heading to, as depicted by the surface plots in Fig. \ref{fig:architecture}. They consist of $H$ stacked vectors obtained similarily to $p_{lat}$ and $p_{lon}$, respectively, for the next $H$ hectometers of the river section the vessel is heading to, assuming a constant discharge and the continuity of the vessel on the current lane.  
The vessel (dis)location vector $(x_t,y_t,a_t)^T$, and the vectors $p_{lat}$ and $p_{lon}$ are passed through a linear layer each, as shown in Fig. \ref{fig:architecture}. The matrices representing the distribution information of the river section ahead are passed each through a recurrent layer (such as an LSTM or GRU). The outputs of the linear and recurrent layers are concatenated ($\otimes$) to obtain the embedding vectors for each sequence element. The embedded sequence is then split (\rotatebox[origin=c]{90}{$\ominus$}) and the observation part for the time steps up to $t_{obs}-1$ passed to the transformer encoder, while the embedding of the last observation sequence element is passed, together with the remaining time series, i.e. the prediction sequence, to the decoder. 
Thus, the last observation sequence element is used to initialize the transformer decoder. The decoder input is masked to only provide the known information at each prediction time step as described in \cite{Vaswani.2017}. 
Before being processed by the transformer, positional encodings are added ($\oplus$) to the sequence elements to provide the model with information of the position of each element in the time series. As $y_t,  \forall t \in \lbrace 1,...,N \rbrace$ is the displacement in KM per minute, $(x_t,y_t)$ do not contain sufficient positional information to eliminate the need for positional encodings here.  Similarly to \cite{Vemula.2018}, it is assumed, that the lateral and longitudinal trajectory features follow a bivariate normal distribution. Thus, the output at time step $t \in \lbrace t_{obs}+1,...,N \rbrace$, $o_t$, is a 5D vector containing the means, standard deviations, and correlation of a bivariate Gaussian distribution, $(\mu^x_{t}, \mu^y_{t}, \sigma_{t}^x, \sigma^y_{t}, \rho_t)$. The transformer model is trained to minimize the negative log-likelihood loss of the true future vessel (dis)location features $(x_t,y_t)$ under the predicted bivariate Gaussian distribution, for all predicted time steps. 

\section{Data}\label{sec:Data}
Three sections of the Rhine river with straight, slightly curved, and sharply curved sections are covered in this study (KM 556.5-562, 571-579, and 598.5-608.5).  
For generating the GMMs, AIS data of the year 2020 are used, and for training the transformer model, AIS data of 2021. Both datasets are split into trips. 
A VTP model for upstream navigating vessels is developed, thus downstream trips are discarded. Trips used for training and testing the model are interpolated to 1 min time intervals using Cubic Hermite spline interpolation and fairway geometries (center curvatures and curve orientations) are added. All trips are labeled with KMs, KM distances per minute, fairway center offsets (negative for positions left of the center), and discharge values.  
The KMs are determined as in \cite{Donandt.2023}. 
Discharge values are added from measurements of the gauges covering the region of the vessel positions. The considered river sections do not include transitions between regions covered by different gauges. 
Instationarities between the gauge and the vessel location are considered neglectable in this study due to the short time horizon and consequently limited KM range. The discharge measurements, available at each 15 min, are provided by the German Federal Waterways and Shipping Agency\footnote{\url{https://www.gdws.wsv.bund.de}} and the German Federal Institute of Hydrology\footnote{\url{https://www.bafg.de}}. The AIS data, trip splitting functionality, fairway geometries, and information of the delimitations of the river sections for each gauge are provided by the BAW. For training the model, the labeled trips are shuffled and split into training, validation, and test dataset in a 80-10-10 ratio, and sequences of $N$ time steps are sampled from each trip without overlapping. The total number of sequences obtained is 264.272.  
The distribution feature vectors $p_{lat}$ and $p_{lon}$ are dynamically added to the sequences during training/validation/testing by using $L_{lat}$ and $L_{lon}$. Lookup failures are handeled by shifting the requested discharge and/or lane up to a tolerance threshhold of +/-500 m³/s and +/-1 lane. Any outliers beyond these thresholds are discarded.  

\begin{table}[]
	\centering
	\caption{Training duration and model complexity}\label{tab:complexity}
	\begin{tabular}{m{0.14\textwidth}m{0.07\textwidth}m{0.09\textwidth}m{0.09\textwidth}} 
		\hline
		\textbf{Model} & \textbf{\begin{tabular}[c]{@{}l@{}}Time per \\ epoch {[}s{]}\end{tabular}} & \textbf{\begin{tabular}[c]{@{}l@{}}Last \\ checkpoint\end{tabular}} & \textbf{\begin{tabular}[c]{@{}l@{}}Trainable \\ parameters\end{tabular}}                                                               \\ \hline \hline
		N-CSCT         & 160                                                                        & 926                                                             & 745.418                                                                    \\ \hline
		Trans          & 190                                                                        & 995                                                             & 
		80.039                                                                                                                                                                                                              \\ \hline
		GMM-Trans      & 320                                                                        & 1000                                                            &                                                                                                                                                                        322.215                                   \\ \hline
		GMM-Trans-LSTM & 870                                                                        & 977                                                             & 703.207                                                                                                                                                                                                              \\ \hline
		GMM-Trans-GRU  & 790                                                                        & 981                                                             & 690.407                                                                                                                                                                                                             \\ \hline
	\end{tabular}
\end{table}
\begin{table}[]\centering
	\caption{ADE and FDE after 5 minutes.}
	\label{tab:adefde}
	\begin{tabular}{m{0.16\textwidth}m{0.12\textwidth}m{0.12\textwidth}}
		\hline
		\textbf{Model}      & \textbf{ADE}          & \textbf{FDE}          \\ \hline \hline
		Const-Acc & 41.38$\pm$28.75          & 81.98$\pm$61.4           \\ \hline
		Const-Vel     & 24.91$\pm$16.07          & 43.48$\pm$30.52          \\ \hline
		N-CSCT \cite{Donandt.2023}             & 18.67$\pm$16.09          & 30.10$\pm$29.65           \\ \hline
		TP-Baseline         & 29.25$\pm$21.68          & 43.23$\pm$35.58          \\ \hline
		GMM-Baseline        & 29.71$\pm$22.26          & 48.30$\pm$41.85           \\ \hline
		Trans               & 18.03$\pm$13.05          & 29.34$\pm$25.53          \\ \hline
		GMM-Trans           & 15.14$\pm$12.23          & 23.79$\pm$23.73          \\ \hline
		GMM-Trans-LSTM      & 14.61$\pm$12.10           & 23.00$\pm$23.53           \\ \hline
		GMM-Trans-GRU      & \textbf{14.54$\pm$11.94} & \textbf{22.88$\pm$23.35} \\ \hline
	\end{tabular}
\end{table}
\section{Results}\label{sec:Results}
The prediction performance of the proposed model is compared to several non DL-based baselines and to the context-sensitive classification transformer - N-CSCT \cite{Donandt.2023}.  
Additionally, an ablation study is conducted to assess the advantages of incorporating the proposed features which represent the local and discharge-dependent distributions of both lateral offsets from the fairway center and lane-specific KM distances per minute. 
\subsection{Baselines \& benchmark}
Four baselines are employed, ranging from naive extrapolation methods to more sophisticated approaches incorporating statistical evaluations, including the Gaussian Mixture Models fitted in this study: 
\begin{itemize}
		\item Const-Vel: maintains last observed KM distance per minute and fairway center offset constant during prediction;
		\item Const-Acc: replaces last observed KM distance per minute in Const-Vel by its change rate; 
		\item TP-Baseline\cite{Donandt.2023}: holds mean deviation of observed KM distances per minute from typical ones and observed average Euclidean distance from the typical route constant during prediction; see \cite{Donandt.2023}	for details on the generation of the typical route and typical KM distance per minute;
		\item GMM-Baseline: holds the distance between the last observed fairway center offset and the closest fairway center offset with the maximum density value in the corresponding $p_{lat}$ and  the difference between the last observed KM distance per minute and the closest KM distance per minute with maximum density value in the corresponding $p_{lon}$ constant during prediction.
\end{itemize}
The 
N-CSCT \cite{Donandt.2023} is used as a benchmark model. It makes use of statistical pre-evaluation results as well, but instead of appending statistical information to the input feature space, the typical route and typical KM distances per minute, also used by TP-Baseline, are utilized to define the lateral and longitudinal dislocation features of the input and output sequences. Additionally, river curvature and curve orientation are used to initialize the decoder \cite{Donandt.2022}. The N-CSCT is trained, validated, and tested on the same data source as the proposed model to ensure comparability. 

In this study, the lateral features (fairway center offset) is not a dislocation feature (different from N-CSCT), while the longitudinal feature (KM distance per minute) is. 
As the longitudinal dislocation is calculated in the forward direction, the KM of the prediction time step $t$ is obtained from the predicted KM distance per minute of the previous time step $t-1$. For the first prediction, the KM is consequently already known from the last observed KM distance per minute. To ensure comparability with the baselines and the N-CSCT model, the first predicted KM 
of these models is replaced by the true one.
\begin{figure*}[htb!]
	\includegraphics[trim= 95 0 105 0,clip, width=\textwidth]{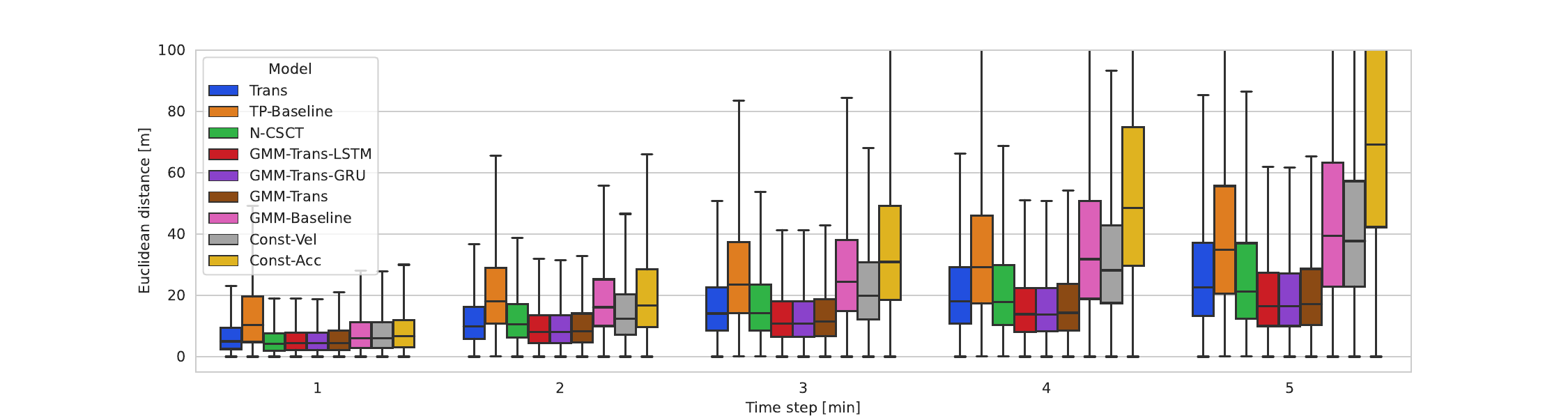}
	\caption{Prediction error per time step of proposed model variants and baselines.}
	\label{fig:errors}
\end{figure*}
\subsection{Model variants of ablation study}
Four model variants are examined, each differing in the input features considered and/or the type of recurrent neural network (RNN) utilized (if applicable):
\begin{itemize}
	\item Trans: only the vessel (dis)location vectors $(x_t,y_t,a_t)^T$, $t \in \lbrace 1,...,t_{obs}\rbrace$ are used as input;
	\item GMM-Trans: $p_{lat}$ and $p_{lon}$ vectors are processed in addition to the vessel (dis)location vectors;
	\item GMM-Trans-LSTM: considers all proposed input feature vectors and matrices, and the matrices are processed by an LSTM;
	\item GMM-Trans-GRU: same as GMM-Trans-LSTM but matrix processing is realized with a GRU.
\end{itemize}

\subsection{Implementation details}
All models are trained for up to 1000 epochs with a batch size of 128 and 16 dataloader workers. The training is performed on CPUs of type AMD EPYC 7742 (Rome) with 64 cores and a clock rate of 2.25 GHz. Insight into the training duration and complexity of the proposed model, its variants, and the N-CSCT model are given in Table  \ref{tab:complexity}. The higher number of trainable parameters in N-CSCT  stems from the classification re-framing of the regression task leading to high-dimensional inputs, and a complexer transformer architecture. 
While the proposed model has one attention head, a single encoder and decoder layer, and a feedforward layer size of 512, the N-CSCT has two attention heads, two encoder and decoder layers, and a feedforward layer size of 1024. 

\subsection{Prediction performance}
Five time steps (i.e. 5 min) are predicted based on an observation sequence of 6 time steps (the last one used for decoder initialization (see Sec. \ref{sec:Model})). While the N-CSCT processes only 5 time steps as input, they correspond to 6 actual positions since the lateral and longitudinal displacements are given as deltas to the next time step.   
The prediction error of a sample is measured by the average displacement error (ADE) calculated by taking the average of the Euclidean distances between predicted and actual positions over all predicted time steps, and the final  displacement error (FDE), only considering the Euclidean distance between prediction and target at the last prediction time step.   
Given that the output of the proposed model comprises the distribution parameters of a bivariate Gaussian distribution rather than a specific position, the final prediction sequence is obtained by sampling 40 pairs of fairway center offsets and KM distances per minute per prediction step from the obtained distribution and taking the mean as the actual prediction. 
In Table \ref{tab:adefde}, the ADE and FDE with standard deviation are reported for all model variants, baselines, and the benchmark model. In Figure \ref{fig:errors}, the boxplots illustrate the distribution of the FDE across increasing prediction horizons. 
The best performing model is the GMM-Trans-GRU, suggesting that the integration of all GMM-based features enhances prediction accuracy. While N-CSCT and Trans exhibit similar means, the N-CSCT's standard deviations are notably higher. Surprisingly, the more sophisticated baselines are surpassed by Const-Vel. Furthermore, TP-Baseline outperforms GMM-Baseline, despite relying on a single typical route and a single typical speed per hectometer, overlooking multi-modality and speed variations along the fairway width. Overall, the proposed model and its variants demonstrate superior performance in terms of ADE and FDE for prediction lengths $>1$. For the first prediction time step, N-CSCT performs marginally better, as depicted in  Fig. \ref{fig:errors}. 

\section{Conclusion}
The results of this study show that the integration of GMM-derived feature vectors, which represent local and discharge-dependent distributions of fairway center offsets and waterway kilometer distances per minute, leads to reduced prediction errors in the comparison. The comparative study is characterized by the omission of this distribution information or the exclusive consideration of a single typical route and typical speeds per hectometer, which - as the result demonstrates - oversimplifies the actual complexity of the navigation behavior of inland vessels. Best outcomes are observed when distribution information from both the current vessel location and the upcoming river section are considered. 
Future works should address several challenges. 
First, a model architecture and hyperparameter optimization is required. A preliminary hyperparameter test, not detailed in this contribution, showed that optimization potential exists when increasing the model complexity. 
Architectural changes, such as consistently using transformers for all sequence encoding tasks or replacing RNNs with CNNs or GNNs to encode the ``maps'' derived from stacked vectors of lateral and longitudinal distribution information, need to be investigated. Also, the superiority of transformers over RNNs in the given sequence-to-sequence task needs to be demonstrated. 
Additionally, the scope of discharge data needs to be expanded to include more variations, and variations of the discharge at river confluences need to be taken into account. Furthermore, individual vessel characteristics, such as width, length, and draught, significantly impact the driving behavior and should be considered. As these values are manually added to the AIS data, their inclusion is not trivial, necessitating reliability checks in the data preprocessing stage. Other relevant AIS features that could improve the prediction quality include heading and maneuver indicator. However, these features are often missing due to the absence of required sensors. 
One major limitation is the unawareness of the traffic situation. Nearby vessels might have a strong influence on the navigation patterns. Thus, the usage of fine-grained spatio-hydrological distribution information in a joint trajectory prediction models such as the one proposed in \cite{Liu.2024} and \cite{Wang.2021pred}, which are becoming increasingly relevant especially for the inland domain, is therefore planned. It is expected that the proposed distribution information can help a joint trajectory prediction model to better account for the surrounding vessels' intents. By addressing these challenges and limitation in future research, the accuracy and robustness of inland vessel trajectory prediction models can be further improved, ultimately enhancing safety and efficiency in navigational decision-making.

\section*{Acknowledgment}
The authors thank the BAW for the provision of AIS data, river-specific data, and the training infrastructure.

\bibliographystyle{IEEEtran}
\bibliography{sn-bibliography}

\end{document}